%% file: main.tex
\documentclass[runningheads,orivec]{llncs}
\usepackage[T1]{fontenc}
\usepackage{graphicx,verbatim}
\usepackage{booktabs}
\usepackage{amsmath}
\usepackage{amssymb}
\usepackage{rotating} 
\usepackage{multirow} 
\usepackage{placeins}
\usepackage{hyperref}

\begin{document}
%
\title{Prototype-Based Knowledge Guidance for Fine-Grained Structured Radiology Reporting}
\titlerunning{ProtoSR}

\author{Chantal Pellegrini\thanks{Shared first authorship.}\inst{1,2}, Adrian Delchev\inst{\star,1}, Ege Özsoy\inst{1,2}, Nassir Navab\inst{1,2}, Matthias Keicher\inst{1,2}}

\institute{Computer Aided Medical Procedures, Technische Universit{\"a}t M{\"u}nchen, Germany\\
\and Munich Center for Machine Learning, Germany}

\maketitle              
\begin{abstract}
Structured radiology reporting promises faster, more consistent communication than free text, but automation remains difficult as models must make many fine-grained, discrete decisions about rare findings and attributes from limited structured supervision. In contrast, free-text reports are produced at scale in routine care and implicitly encode fine-grained, image-linked information through detailed descriptions. To leverage this unstructured knowledge, we propose ProtoSR, an approach for injecting free-text information into structured report population. First, we introduce an automatic extraction pipeline that uses an instruction-tuned LLM to mine 80k+ MIMIC-CXR studies and build a multimodal knowledge base aligned with a structured reporting template, representing each answer option with a visual prototype. Using this knowledge base, ProtoSR is trained to retrieve prototypes relevant for the current image-question pair and augment the model predictions through a prototype-conditioned residual, providing a data-driven second opinion that selectively corrects predictions. On the Rad-ReStruct benchmark, ProtoSR achieves state-of-the-art results, with the largest improvements on detailed attribute questions, demonstrating the value of integrating free-text derived signal for fine-grained image understanding.


\end{abstract}

\input{chapters/introduction}
\input{chapters/method}
\input{chapters/experiments}

\input{chapters/conclusion}

%
%
%

\bibliographystyle{splncs04}
\bibliography{refs}

\end{document}

%% file: chapters/introduction.tex
\section{Introduction}

Medical imaging is central to diagnosis and clinical decision making, and radiologists communicate findings primarily through radiology reports, typically written in free text. However, free-text reporting is time-consuming and often non-standardized, leading to variability in completeness, and clarity \cite{pino2021clinically,hong2013content,pellegrini2023rad}. In contrast, structured radiology reporting organizes findings into predefined fields and standardized answer options, rather than relying solely on unconstrained narrative text. By using controlled vocabularies and templates, it can improve consistency and completeness while also enabling secondary uses such as quality monitoring and downstream analysis  \cite{hong2013content,nobel2022structured,jorg2023implementation,pellegrini2023rad}.

With growing demands in radiology, automated radiology report generation has the potential to support reporting workflows, but most work has focused on free-text generation \cite{wang2025cxpmrg,bannur2024maira,pellegrini2025radialog,xiao2025radiology,lee2025cxr}. In parallel, generalist medical vision--language models such as MedGemma and CheXagent unify multiple radiology tasks, including report writing and question answering, within a single framework \cite{sellergren2025medgemma,chexagent-2024}. Automated structured reporting (SR) is comparatively less explored: earlier work approached it via prompt-based template population or disease/attribute-specific classifiers \cite{pino2021clinically,keicher2022few,kale2023replace,bhalodia2021improving}. Rad-ReStruct \cite{pellegrini2023rad} introduced a fine-grained SR benchmark with a hierarchical template spanning findings and detailed attributes, and proposed an iterative VQA-style model to populate predefined fields. Context-VQA \cite{arsalane2024context} further improved structured-report population on Rad-ReStruct by incorporating report-derived context during training.

A remaining challenge in automated structured reporting is that fine-grained templates include many rare attributes \cite{pellegrini2023rad}, while structured datasets such as Rad-ReStruct are limited in size, providing sparse supervision. On the other hand, large public datasets such as MIMIC-CXR \cite{Johnson2019mimic} provide hundreds of thousands of paired chest radiographs and free-text reports, offering broad coverage across common and rare findings. However, these reports differ significantly in style and word choice, complicating direct mapping into a strict SR taxonomy. Recent instruction-tuned LLMs make it increasingly feasible to distill free-text reports into standardized, template-aligned signals \cite{woznicki2025automatic,delbrouck2025automated}. This opens the possibility of using large-scale free-text collections as an auxiliary knowledge source to improve image-based structured reporting. Knowledge integration has been explored in free-text reporting in the form of feature-level fusion of e.g. prior patient information or metadata \cite{zhu2023utilizing,bannur2024maira} and via retrieval of similar cases using prototype memories or knowledge bases \cite{wang2022cross,yang2023radiology,zhang2025historical,sun2025fact}. RadIR \cite{zhang2025radir} mines fine-grained supervision from free-text radiology reports for scalable retrieval, but does not address how retrieved evidence can be injected into structured prediction pipelines. Overall, these approaches mainly operate in an unstructured output space, whereas SR requires mechanisms that influence fine-grained discrete decisions.\looseness=-1

To close this gap, we propose ProtoSR, a prototype-conditioned late-fusion framework for fine-grained structured radiology reporting that leverages information extracted from routine free-text reports. ProtoSR transforms paired images and free-text reports into an explicit prototype memory and learns to retrieve prototypes that influence discrete per-field answer selection, enabling targeted corrections of long-tail decisions. We make two main contributions: we propose an LLM-driven mining, normalization, and filtering pipeline that converts a large-scale free-text report collection into a multimodal prototype knowledge base aligned with a structured reporting template and a prototype-conditioned late-fusion module that converts retrieved examples into an answer-aligned correction signal, selectively revising fine-grained predictions while preserving the base model’s overall behavior. Our experiments demonstrate consistent gains, with the strongest improvements on detailed attribute decisions, indicating that routine free-text reports can be leveraged as knowledge signal to improve fine-grained understanding needed for structured reporting.

%% file: chapters/method.tex
\section{Method}

\begin{figure}[tb]
  \centering
  \includegraphics[width=0.96\linewidth]{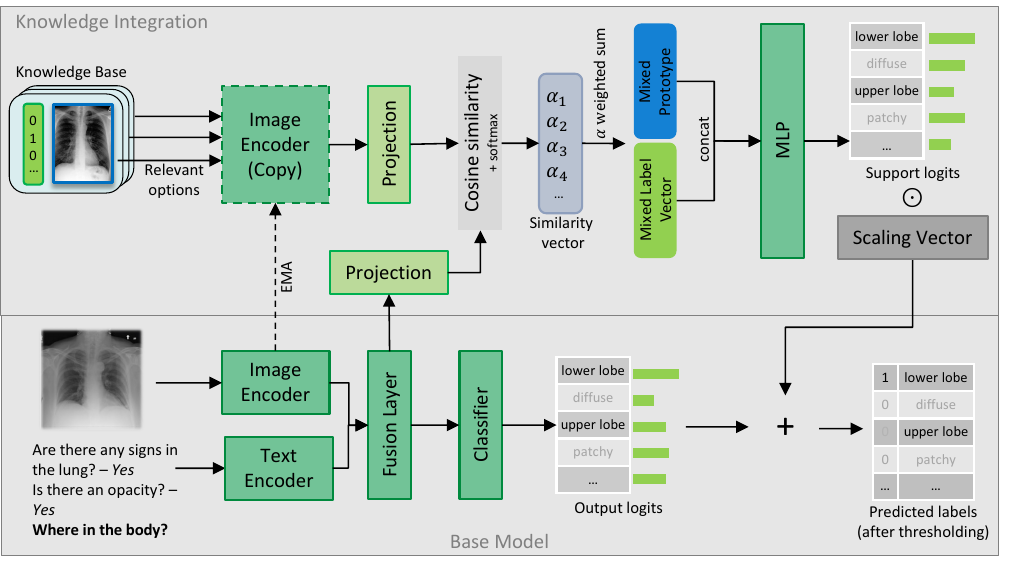}
    \caption{Overview of our architecture. The hierarchical SR base model produces base logits, while the prototype-conditioned knowledge branch retrieves label-aligned examples from a prototype bank and converts them into a scaled static residual logit correction. The final prediction is made from the fused logits.}
  \label{fig:architecture}
\end{figure}

Structured reporting datasets are often small and imbalanced, with studies containing diverse fine-grained attributes (e.g., location, visual appearance, severity) being underrepresented. To address this we propose ProtoSR, a structured reporting method, which can leverage knowledge from large free-text report databases. First, we mine free-text radiology reports to recover label-specific visual prototypes aligned to a fixed, fine-grained structured reporting template. We then leverage this knowledge base to retrieve prototype evidence and inject it into a hierarchical structured-reporting vision-language model.

\subsection{Knowledge Base Construction}
\begin{figure}[tb]
  \centering
  \includegraphics[width=0.96\linewidth]{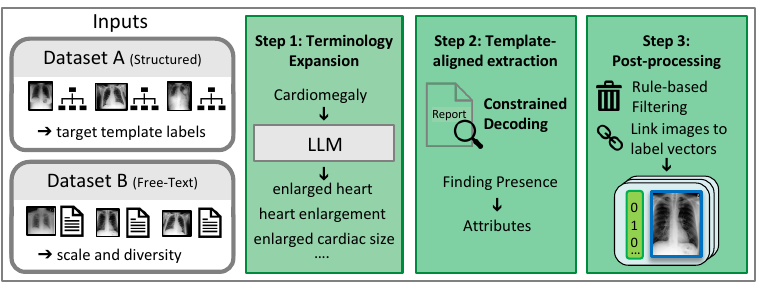}
  \caption{Knowledge base extraction. Dataset A defines the target structured reporting template, while Dataset B contains paired images and free-text reports. We first expand the template label vocabulary with alternative phrasings, then identify template-aligned label occurrences in the free-text reports, and finally apply filtering to build a knowledge base that links Dataset B images to the template-aligned labels.}
  \label{fig:kb}
\end{figure}

We construct a template-aligned prototype knowledge base by mapping a large corpus of paired images and free-text reports (dataset $B$) to the fine-grained label space defined by a structured dataset (dataset $A$). Dataset $A$ defines the target template with each template answer option corresponding to a label $\ell \in L$. Exploiting the report--image pairs in $B$, any study whose report affirms $\ell$ is treated as a candidate exemplar for $\ell$. The knowledge base construction process shown in Fig. \ref{fig:kb} consists of the following steps:\\
\noindent \textbf{Terminology expansion:}
To improve robustness to reporting variability in $B$, we expand the vocabulary associated with each target label $\ell$. We leverage a zero-shot LLM to propose synonyms, abbreviations, and alternative phrasings, yielding a dictionary that maps description variants to the canonical label.\\
\textbf{Template-constrained extraction:}
For each report $r \in B$, we extract template-aligned labels. We first query the LLM to decide whether a finding corresponding to $\ell$ is contained in a report. If present, a subsequent query extracts the corresponding attribute value(s) specified by the template. We apply the extraction hierarchically, querying deeper attributes only when their parent finding is present and use constrained decoding, where only template-aligned answers are kept, to ensure valid outputs.\\
\textbf{Post-processing and Knowledge Base assembly:}
We apply rule-based filters to reduce noise and enforce consistency with 
\(A\)’s ontology. Uncertain or invalid extractions are discarded, and hierarchical constraints are enforced by removing positive parent labels when none of their child labels are supported. Each retained tuple is then linked to its imaging study, yielding an example pool for each $\ell \in L$. To construct the prototype bank, we uniformly sample up to $K$ images from each label-specific example pool and aggregate their image-encoder embeddings into a single prototype using element-wise max pooling, preserving the strongest signals across the sampled images.

\subsection{Knowledge-Enhanced Late Fusion Architecture}
Given the current image and question, we retrieve prototypes from the knowledge base and convert their evidence into scaled logit corrections that are added to the base model logits. Fig.~\ref{fig:architecture} shows an overview of the proposed architecture.

\noindent\textbf{Base model:}
Our structured reporting backbone follows the architecture of Rad-ReStruct \cite{pellegrini2023rad}. The model takes an image $x$ and a question context $q$, consisting of the current question and prior question--answer pairs. Pretrained image and text encoders extract features that are fused by a transformer module,
\[
S = f_{\text{fusion}}\!\bigl(f_{\text{img}}(x),\, f_{\text{txt}}(q)\bigr),
\]
and a classifier head predicts logits over the answer space $Y$,
\[
z_{\text{base}} = f_{\text{cls}}(S) \in \mathbb{R}^{|Y|}.
\]
\\
\textbf{Prototype-conditioned knowledge branch:}
We propose to augment the backbone with a prototype-conditioned knowledge branch that retrieves visually similar examples from the mined knowledge base and converts them into an answer-aligned support bias for the current decision. Concretely, given the question-conditioned fused representation $S$, we retrieve evidence from the knowledge base containing $M$ prototype embeddings $P\in\mathbb{R}^{M\times d}$ with hidden dimension $d$, and the associated answer vectors $A\in\mathbb{R}^{M\times |Y|}$, where each row is a one-hot encoding for the prototype’s label. To do so, we project $S$ and prototype embeddings to a shared space using a linear projection layer and compute cosine-similarity weights $\alpha \in\mathbb{R}^{M}$ between the projected representations, , considering only prototypes whose labels correspond to valid answer options for the current question. Retrieved evidence is summarized as a prototype feature vector and an answer vector:
\[
v=\alpha^\top P \in \mathbb{R}^{d},
\qquad
u=\alpha^\top A \in \mathbb{R}^{|Y|}
\]
Here, $v$ summarizes retrieved visual evidence as a weighted average in the prototype-embedding space, while $u$ aggregates the corresponding one-hot prototype labels into a support vector with soft scores for all labels in the answer space $Y$. These two vectors are then concatenated and transformed to a support bias $b_{\text{sup}}$ using an MLP, which learns to predict how the combination of ``visual similarity'' and ``answer tendency'' should adjust the output scores:
\[
b_{\text{sup}}=\mathrm{MLP}([v;u])\in\mathbb{R}^{|Y|}
\]
If no compatible prototype exists, we set $b_{\text{sup}}=\mathbf{0}$
\\
\textbf{Late fusion and training:}
Our final logits are obtained by combining the backbone prediction with the knowledge-derived bias via a learned scaling vector $s\in\mathbb{R}^{|Y|}$, which calibrates the influence of retrieved evidence per answer dimension:
\[
z_{\text{final}} = z_{\text{base}} + s \odot b_{\text{sup}}
\]
This late-fusion design preserves the backbone decision pathway and enables targeted corrections where prototype evidence is informative. Further, the knowledge branch is lightweight, only adding an MLP and a per-answer scaling vector. We train the full model end-to-end using the same multi-label objective as Rad-ReStruct, applied to $z_{\text{final}}$. The prototype bank is treated as external memory, with periodic updates of prototype vectors using the current image encoder to maintain alignment with the continuously fine-tuned encoder. 

%% file: chapters/experiments.tex
\section{Experimental Setup}

\textbf{Datasets:}
We evaluate on Rad-ReStruct \cite{pellegrini2023rad}, a fine-grained hierarchical structured reporting benchmark with 3{,}597 chest X-ray studies paired with structured reports. The template is organized into three question levels: Level 1 (L1) queries coarse abnormality existence (25 questions), Level 2 (L2) queries specific findings (216 questions), and Level 3 (L3) queries fine-grained attributes of findings such as location, appearance, and severity (477 questions). We follow the official splits and the evaluation protocol from Rad-ReStruct. For knowledge mining, we use MIMIC-CXR \cite{Johnson2019mimic}, an independently collected dataset that is disjoint from Rad-ReStruct and contains 227{,}835 radiology reports paired with chest X-rays, and mine structured labels from the \emph{Findings} and \emph{Impression} sections.\\
\textbf{Implementation Details: }
For structured-label extraction from MIMIC-CXR reports, we use Qwen2.5-7B-Instruct \cite{qwen25}.  To keep prototype embeddings aligned with the continuously fine-tuned encoder, we compute prototype embeddings using an exponential moving average (EMA) copy of the image encoder and refresh the knowledge base every 10k training steps. We aggregate up to $K=5$ images for each prototype (fewer if not available). We follow Rad-ReStruct \cite{pellegrini2023rad} for the structured reporting (SR) backbone, using EfficientNet-B5 \cite{tan2019efficientnet} as image encoder and RadBERT \cite{yan2022radbert} as text encoder. We train for 34 epochs with Adam optimizer, learning rate $1\mathrm{e}{-5}$, batch size 8, and gradient accumulation 4 on one Nvidia RTX 3090 GPU with 24 GB memory. Evaluation follows the evaluation procedure used in Rad-ReStruct, iteratively querying the full template while appending prior question--answer pairs to the context. For baselines not designed for multi-turn dialogue (e.g., MedGemma, CheXagent), we compress the history into a single prompt summary followed by the current question, and include the answer options to enforce selection from the constrained answer space.

\section{Results and Discussion}

\subsection{Quality and coverage of knowledge base mining}
Our knowledge base requires mining template-aligned labels from free-text reports so that mined studies can be indexed into answer-option prototypes. Since the target free-text corpus (MIMIC-CXR) does not provide ground-truth answers in the Rad-ReStruct template space, extraction correctness cannot be measured directly. We therefore evaluate our extraction pipeline on Rad-ReStruct, which provides structured labels linked to free-text reports, providing corresponding ground-truth template answers, and compare several instruction-tuned LLM backbones for extraction.

\begin{table}[tb]
  \centering
  \caption{Evaluation of candidate instruction-tuned LLMs with and without terminology expansion for template-aligned extraction on Rad-ReStruct.}
  \label{tab:llm-extraction-quality}
  \setlength{\tabcolsep}{5pt}
  \begin{tabular}{lcccccc}
    \toprule
    & \multicolumn{2}{c}{L1-F$_1$} & \multicolumn{2}{c}{L2-F$_1$} & \multicolumn{2}{c}{L3-F$_1$} \\
    \cmidrule(lr){2-3} \cmidrule(lr){4-5} \cmidrule(lr){6-7}
    Terminology Expansion & X & \checkmark & X & \checkmark & X & \checkmark \\
    \midrule
    Llama 3.1 8B Instruct & \textbf{76.4} & 85.2 & 72.5 & 86.9 & 65.7 & 74.0 \\
    Mistral 8B Instruct   & 71.7 & 78.0 & \textbf{72.8} & 75.4 & \textbf{69.4} & 77.0 \\
    Qwen2.5 7B Instruct   & 72.5 & \textbf{86.8} & 72.2 & \textbf{87.4} & 68.1 & \textbf{80.6} \\
    \bottomrule
  \end{tabular}
\end{table}

Given a free-text report, the extractor predicts the answer option for each template question at hierarchy levels L1 to L3. We compare three instruction-tuned LLMs (Mistral 8B, Llama 3.1 8B, Qwen2.5 7B) with and without terminology expansion, evaluating macro-F$_1$ as defined in Rad-ReStruct \cite{pellegrini2023rad}. For all models, the proposed terminology expansion clearly improves extraction quality on all levels, highlighting the importance of handling clinical paraphrases, synonyms and abbreviations. Overall, Qwen2.5-7B-Instruct with terminology expansion achieves the strongest correctness results, and is therefore used in our final pipeline.

Using the resulting extraction pipeline, we mine MIMIC-CXR and obtain broad label-space coverage after post-processing: 100\% at L1, 96\% at L2, and 82\% at L3 (see Table~\ref{tab:kb-coverage}). This yields substantial prototype support even for fine-grained attributes, which are typically long-tailed in supervised SR datasets.

\begin{table}[tb]
  \centering
  \caption{Knowledge base coverage across hierarchy levels after post-processing. A category is counted as covered if at least one mined instance is assigned to it.}
  \label{tab:kb-coverage}
  \begin{tabular}{lccc}
    \toprule
    Level & Total categories & Covered categories & Coverage \\
    \midrule
    L1 & 56 & 56 & 100\% \\
    L2 & 326 & 314 & 96\% \\
    L3 & 1167 & 966 & 82\% \\
    \bottomrule
  \end{tabular}
\end{table}

\subsection{Structured Reporting Performance}

Table~\ref{tab:main} reports performance on Rad-ReStruct. We compare against methods that predict from images without report-derived context at inference. We exclude HiCA-VQA \cite{zhang2025hierarchical}, which uses LLM summaries of the structured reports as input, representing a different task formulation. General-purpose medical VLLMs, such as MedGemma and CheXagent, achieve competitive overall scores, but still lag behind specialized structured-reporting models. This gap suggests that Rad-ReStruct benefits from architectures and training objectives tailored to hierarchical structured reporting and fine-grained, field-level supervision. Compared to both generalist medical VLLMs and prior structured reporting methods, ProtoSR achieves the best overall F$_1$ and the strongest performance at the deeper levels of the hierarchy, with the largest gains on the fine-grained attribute questions (L3), where supervision is most sparse and errors are most frequent, highlighting the value of prototype guidance for long-tail attribute configurations.

\begin{table}[tbh]
  \centering
  \caption{Performance on the Rad-ReStruct benchmark.}
  \label{tab:main}
  \begin{tabular}{lccccc}
    \toprule
    Method & Overall F$_1$ & L1-F$_1$ & L2-F$_1$ & L3-F$_1$ & Report Acc. \\
    \midrule
    MedGemma \cite{sellergren2025medgemma} & 26.8 & 38.2 & 63.4 & 2.8 & 0.0\%\\
    CheXagent \cite{chexagent-2024} & 32.4 & 62.1 & 69.8 & \underline{6.2} & 20.3\%\\
    RaDialog \cite{pellegrini2025radialog} & 28.7 & 56.8 & 70.0 & 0.23 & \underline{39.6\%}\\
    hi-VQA \cite{pellegrini2023rad} & 32.0 & 64.6 & 71.6 & 4.1 & 32.6\%\\
    Context-VQA \cite{arsalane2024context} & \underline{32.9} & \textbf{67.2} & \underline{71.8} & 3.2  & \textbf{39.7\%}\\
    ProtoSR & \textbf{34.4} & \underline{66.2} & \textbf{72.8} & \textbf{7.4} & 36.6\% \\
    \bottomrule
  \end{tabular}
\end{table}

While some methods outperform in report-level accuracy, this metric provides a complementary but imperfect view of performance. It can favor conservative strategies that answer only a few fields or default to ``no finding'', increasing exact-match scores on normal studies but harming performance on abnormal cases. This can be observed for RaDialog and Context-VQA, which achieve high report accuracy despite lower F$_1$, suggesting that partial or empty reports are favored.\looseness=-1

To isolate the effect of prototype guidance, we compare our model against the same base model without knowledge integration. As shown in Table~\ref{tab:ablations}, adding prototype-guided late fusion yields consistent gains across levels, with the most pronounced improvement at L3 with a relative improvement of +72.1\%. Report-level accuracy also increases, indicating improved end-to-end consistency for our backbone.  We further ablate the knowledge integration strategy in Table~\ref{tab:ablations} comparing our method to an early-fusion variant that includes a list of all answer options paired with their knowledge embeddings directly into the input sequence. We see that this model can not utilize the knowledge effectively, leading to almost unchanged performance compared to the base model. Finally, to verify that improvements come from the content of retrieved prototypes rather than added fusion capacity, we perform an ablation that replaces prototypes with Gaussian noise, while keeping the architecture unchanged. The performance falls back to baseline level, suggesting ProtoSR exploits meaningful prototype structure and learns to ignore uninformative signals.

\begin{table}[tb] 
  \centering
  \caption{Ablations across knowledge-integration strategies.}
  \label{tab:ablations}
  \begin{tabular}{lcccc}
    \toprule
    Variant & Overall F1 & L1-F$_1$ & L2-F$_1$ & L3-F$_1$ \\
    \midrule
    No knowledge & 32.5 & 64.2 & 71.3 & 4.3 \\
    Early Fusion &  32.5 & 64.8 & 71.4 & 4.3 \\
    Randomized prototypes & 32.7 & 64.3 & 71.4 & 4.4 \\
    ProtoSR & \textbf{34.4} & \textbf{66.2} & \textbf{72.8 }& \textbf{7.4} \\
    \bottomrule
  \end{tabular}
\end{table}

%% file: chapters/conclusion.tex
\FloatBarrier
\section{Conclusion}
In this work, we propose ProtoSR, a knowledge-guided approach to structured radiology reporting that aligns information mined from abundant free-text reports with a fine-grained structured template and integrates it into structured report population. Our template-aligned extraction pipeline converts free-text reports into label-linked example prototypes, and our prototype-conditioned structured reporting model that leverages exemplars during prediction. By injecting retrieved evidence as a residual at the logit level, our method targets the long tail of detailed attribute configurations while preserving the strengths of the hierarchical structured reporting backbone. Overall, our results indicate that routine free-text reports can be systematically converted into template-aligned, image-linked knowledge that improves fine-grained structured radiology reporting.

\section*{Acknowledgements}
The authors gratefully acknowledge the financial support by the Bavarian Ministry of Economic Affairs, Regional Development and Energy (StMWi) under project ThoraXAI (DIK-2302-0002), and the German Research Foundation (DFG, grant 469106425 - NA 620/51-1).

%% file: refs.bib
@inproceedings{pellegrini2023rad,
  title={Rad-restruct: A novel vqa benchmark and method for structured radiology reporting},
  author={Pellegrini, Chantal and Keicher, Matthias and {\"O}zsoy, Ege and Navab, Nassir},
  booktitle={MICCAI},
  pages={409--419},
  year={2023},
  organization={Springer}
}

@article{Johnson2019mimic,
  title={MIMIC-CXR-JPG, a large publicly available database of labeled chest radiographs},
  author={Johnson, Alistair EW and Pollard, Tom J and Greenbaum, Nathaniel R and Lungren, Matthew P and Deng, Chih-ying and Peng, Yifan and Lu, Zhiyong and Mark, Roger G and Berkowitz, Seth J and Horng, Steven},
  journal={arXiv preprint arXiv:1901.07042},
  year={2019}
}

@misc{qwen25,
      title={Qwen2.5 Technical Report}, 
      author={An Yang and Baosong Yang and Beichen Zhang and Binyuan Hui and Bo Zheng and Bowen Yu and Chengyuan Li and others},
      year={2025},
      eprint={2412.15115},
      archivePrefix={arXiv},
      primaryClass={cs.CL},
      url={https://arxiv.org/abs/2412.15115}, 
}

@inproceedings{pino2021clinically,
  title={Clinically correct report generation from chest x-rays using templates},
  author={Pino, Pablo and Parra, Denis and Besa, Cecilia and Lagos, Claudio},
  booktitle={International Workshop on Machine Learning in Medical Imaging},
  pages={654--663},
  year={2021},
  organization={Springer}
}

@article{keicher2022few,
  title={Few-shot structured radiology report generation using natural language prompts},
  author={Keicher, Matthias and Mullakaeva, Kamilia and Czempiel, Tobias and Mach, Kristina and Khakzar, Ashkan and Navab, Nassir},
  journal={arXiv preprint arXiv:2203.15723},
  year={2022}
}

@inproceedings{bhalodia2021improving,
  title={Improving pneumonia localization via cross-attention on medical images and reports},
  author={Bhalodia, Riddhish and Hatamizadeh, Ali and Tam, Leo and Xu, Ziyue and Wang, Xiaosong and Turkbey, Evrim and Xu, Daguang},
  booktitle={MICCAI},
  pages={571--581},
  year={2021},
  organization={Springer}
}

@inproceedings{delbrouck2025automated,
  title={Automated structured radiology report generation},
  author={Delbrouck, Jean-Benoit and Xu, Justin and Moll, Johannes and Thomas, Alois and Chen, Zhihong and Ostmeier, Sophie and Azhar, Asfandyar and Li, Kelvin Zhenghao and Johnston, Andrew and Bluethgen, Christian and others},
  booktitle={ACL},
  pages={26813--26829},
  year={2025}
}

@inproceedings{arsalane2024context,
  title={Context-Guided Medical Visual Question Answering},
  author={Arsalane, Wafa and Chikontwe, Philip and Luna, Miguel and Kang, Myeongkyun and Park, Sang Hyun},
  booktitle={Meets Africa Workshop},
  pages={245--255},
  year={2024},
  organization={Springer}
}

@inproceedings{pellegrini2025radialog,
  title={Radialog: Large vision-language models for x-ray reporting and dialog-driven assistance},
  author={Pellegrini, Chantal and {\"O}zsoy, Ege and Busam, Benjamin and Wiestler, Benedikt and Navab, Nassir and Keicher, Matthias},
  booktitle={Medical Imaging with Deep Learning},
  year={2025}
}

@article{bannur2024maira,
  title={Maira-2: Grounded radiology report generation},
  author={Bannur, Shruthi and Bouzid, Kenza and Castro, Daniel C and Schwaighofer, Anton and Thieme, Anja and Bond-Taylor, Sam and Ilse, Maximilian and P{\'e}rez-Garc{\'\i}a, Fernando and Salvatelli, Valentina and Sharma, Harshita and others},
  journal={arXiv preprint arXiv:2406.04449},
  year={2024}
}

@inproceedings{wang2025cxpmrg,
  title={Cxpmrg-bench: Pre-training and benchmarking for x-ray medical report generation on chexpert plus dataset},
  author={Wang, Xiao and Wang, Fuling and Li, Yuehang and Ma, Qingchuan and Wang, Shiao and Jiang, Bo and Tang, Jin},
  booktitle={Proceedings of the Computer Vision and Pattern Recognition Conference},
  pages={5123--5133},
  year={2025}
}

@article{lee2025cxr,
  title={CXR-LLAVA: a multimodal large language model for interpreting chest X-ray images},
  author={Lee, Seowoo and Youn, Jiwon and Kim, Hyungjin and Kim, Mansu and Yoon, Soon Ho},
  journal={European Radiology},
  pages={1--13},
  year={2025},
  publisher={Springer}
}

@inproceedings{kale2023replace,
  title={Replace and report: NLP assisted radiology report generation},
  author={Kale, Kaveri and Bhattacharyya, Pushpak and Jadhav, Kshitij},
  booktitle={Findings of the Association for Computational Linguistics: ACL 2023},
  pages={10731--10742},
  year={2023}
}

@inproceedings{zhu2023utilizing,
  title={Utilizing longitudinal chest x-rays and reports to pre-fill radiology reports},
  author={Zhu, Qingqing and Mathai, Tejas Sudharshan and Mukherjee, Pritam and Peng, Yifan and Summers, Ronald M and Lu, Zhiyong},
  booktitle={MICCAI},
  pages={189--198},
  year={2023},
  organization={Springer}
}

@inproceedings{xiao2025radiology,
  title={Radiology report generation via multi-objective preference optimization},
  author={Xiao, Ting and Shi, Lei and Liu, Peng and Wang, Zhe and Bai, Chenjia},
  booktitle={Proceedings of the AAAI Conference on Artificial Intelligence},
  volume={39},
  pages={8664--8672},
  year={2025}
}

@article{woznicki2025automatic,
  title={Automatic structuring of radiology reports with on-premise open-source large language models},
  author={Wo{\'z}nicki, Piotr and Laqua, Caroline and Fiku, Ina and Hekalo, Amar and Truhn, Daniel and Engelhardt, Sandy and Kather, Jakob and Foersch, Sebastian and D’Antonoli, Tugba Akinci and Pinto dos Santos, Daniel and others},
  journal={European Radiology},
  volume={35},
  number={4},
  pages={2018--2029},
  year={2025},
  publisher={Springer}
}

@inproceedings{zhang2025historical,
  title={Historical report guided bi-modal concurrent learning for pathology report generation},
  author={Zhang, Ling and Yun, Boxiang and Li, Qingli and Wang, Yan},
  booktitle={MICCAI},
  pages={343--352},
  year={2025},
  organization={Springer}
}

@inproceedings{wang2022cross,
  title={Cross-modal prototype driven network for radiology report generation},
  author={Wang, Jun and Bhalerao, Abhir and He, Yulan},
  booktitle={European Conference on Computer Vision},
  pages={563--579},
  year={2022},
  organization={Springer}
}

@article{nobel2022structured,
  title={Structured reporting in radiology: a systematic review to explore its potential},
  author={Nobel, J Martijn and van Geel, Koos and Robben, Simon GF},
  journal={European radiology},
  volume={32},
  number={4},
  pages={2837--2854},
  year={2022},
  publisher={Springer}
}

@article{jorg2023implementation,
  title={Implementation of structured reporting in clinical routine: a review of 7 years of institutional experience},
  author={Jorg, Tobias and Halfmann, Moritz C and Arnhold, Gordon and Pinto dos Santos, Daniel and Kloeckner, Roman and D{\"u}ber, Christoph and Mildenberger, Peter and Jungmann, Florian and M{\"u}ller, Lukas},
  journal={Insights into Imaging},
  volume={14},
  number={1},
  pages={61},
  year={2023},
  publisher={Springer}
}

@article{hong2013content,
  title={Content analysis of reporting templates and free-text radiology reports},
  author={Hong, Yi and Kahn Jr, Charles E},
  journal={Journal of digital imaging},
  volume={26},
  number={5},
  pages={843--849},
  year={2013},
  publisher={Springer}
}

@article{yang2023radiology,
  title={Radiology report generation with a learned knowledge base and multi-modal alignment},
  author={Yang, Shuxin and Wu, Xian and Ge, Shen and Zheng, Zhuozhao and Zhou, S Kevin and Xiao, Li},
  journal={Medical Image Analysis},
  volume={86},
  pages={102798},
  year={2023},
  publisher={Elsevier}
}

@inproceedings{sun2025fact,
  title={Fact-aware multimodal retrieval augmentation for accurate medical radiology report generation},
  author={Sun, Liwen and Zhao, James Jialun and Han, Wenjing and Xiong, Chenyan},
  booktitle={Proceedings of the 2025 Conference of the Nations of the Americas Chapter of the Association for Computational Linguistics: Human Language Technologies (Volume 1: Long Papers)},
  pages={643--655},
  year={2025}
}

@inproceedings{tan2019efficientnet,
  title={Efficientnet: Rethinking model scaling for convolutional neural networks},
  author={Tan, Mingxing and Le, Quoc},
  booktitle={International conference on machine learning},
  pages={6105--6114},
  year={2019},
  organization={PMLR}
}

@article{yan2022radbert,
  title={RadBERT: Adapting transformer-based language models to radiology},
  author={Yan, An and McAuley, Julian and Lu, Xing and Du, Jiang and Chang, Eric Y and Gentili, Amilcare and Hsu, Chun-Nan},
  journal={Radiology: Artificial Intelligence},
  volume={4},
  number={4},
  pages={e210258},
  year={2022},
  publisher={Radiological Society of North America}
}

@article{chexagent-2024,
  title={CheXagent: Towards a Foundation Model for Chest X-Ray Interpretation},
  author={Chen, Zhihong and Varma, Maya and Delbrouck, Jean-Benoit and Paschali, Magdalini and Blankemeier, Louis and Veen, Dave Van and Valanarasu, Jeya Maria Jose and Youssef, Alaa and Cohen, Joseph Paul and Reis, Eduardo Pontes and Tsai, Emily B. and Johnston, Andrew and Olsen, Cameron and Abraham, Tanishq Mathew and Gatidis, Sergios and Chaudhari, Akshay S and Langlotz, Curtis},
  journal={arXiv preprint arXiv:2401.12208},
  url={https://arxiv.org/abs/2401.12208},
  year={2024}
}

@article{sellergren2025medgemma,
  title={MedGemma Technical Report},
  author={Sellergren, Andrew and Kazemzadeh, Sahar and Jaroensri, Tiam and Kiraly, Atilla and Traverse, Madeleine and Kohlberger, Timo and Xu, Shawn and Jamil, Fayaz and Hughes, Cían and Lau, Charles and others},
  journal={arXiv preprint arXiv:2507.05201},
  year={2025}
}

@inproceedings{zhang2025radir,
  title={RadIR: A Scalable Framework for Multi-grained Medical Image Retrieval via Radiology Report Mining},
  author={Zhang, Tengfei and Zhao, Ziheng and Wu, Chaoyi and Zhou, Xiao and Zhang, Ya and Wang, Yanfeng and Xie, Weidi},
  booktitle={MICCAI},
  pages={508--518},
  year={2025},
  organization={Springer}
}

@article{zhang2025hierarchical,
  title={Hierarchical modeling for medical visual question answering with cross-attention fusion},
  author={Zhang, Junkai and Li, Bin and Zhou, Shoujun},
  journal={Applied Sciences},
  volume={15},
  number={9},
  pages={4712},
  year={2025},
  publisher={MDPI}
}
